\documentclass{article}
\usepackage[final]{neurips_2024}

\usepackage[utf8]{inputenc} % allow utf-8 input
\usepackage[T1]{fontenc}    % use 8-bit T1 fonts
\usepackage{hyperref}       % hyperlinks
\usepackage{url}            % simple URL typesetting
\usepackage{booktabs}       % professional-quality tables
\usepackage{amsfonts}       % blackboard math symbols
\usepackage{nicefrac}       % compact symbols for 1/2, etc.
\usepackage{microtype}      % microtypography
\usepackage{xcolor}         % colors
\usepackage{graphicx}
\usepackage{amsmath}
\usepackage{tabulary}
\usepackage{tabularx}
\usepackage[none]{hyphenat}
\usepackage{enumitem}
\usepackage[export]{adjustbox}
\usepackage{amsmath}
\usepackage{makecell}
\usepackage{pifont}
\newcommand{\cmark}{\ding{51}}%
\newcommand{\xmark}{\ding{55}}%

\graphicspath{ {images/} }

\title{Internalizing ASR with Implicit Chain of Thought for Efficient Speech-to-Speech Conversational LLM}

\author{%
  Robin Shing-Hei Yuen\\
  University of British Columbia\\
  \texttt{robinysh@student.ubc.ca} \\
  \And
  Timothy Tin-Long Tse\\
  University of British Columbia\\
  \texttt{ttse05@student.ubc.ca} \\
  \And
  Jian Zhu\\
  University of British Columbia\\
  \texttt{jian.zhu@ubc.ca} \\
}

\begin{document}

\maketitle

\vspace{-2em}
\begin{abstract}
\vspace{-.5em}
Current Speech LLMs are predominantly trained on extensive ASR and TTS datasets, excelling in tasks related to these domains.
However, their ability to handle direct speech-to-speech conversations remains notably constrained.
We find that Speech LLMs often rely on an ASR-to-TTS chain-of-thought pipeline (A-T-T-A chain) to generate good responses. The pipeline first recognizes speech into text and generates corresponding text responses before generating speech responses, which introduces significant latency.
We propose a method that implicitly internalizes ASR chain of thought into a Speech LLM (A-T-A chain), allowing it to bypass the ASR transcript generation but still maintain speech conversation capabilities.
Our approach reduces latency and improves the model’s native understanding of speech,
paving the way for more efficient and natural real-time audio interactions. We also release a large-scale synthetic conversational dataset to facilitate further research.
\end{abstract}

\vspace{-1.5em}

\section{Introduction}
\vspace{-.5em}
Pretrained large speech-language models (Speech LLMs) \citep[e.g.,][]{zhang2023speechgpt,zhang2024speechgpt,zhan2024anygpt,chu2023qwen,chu2024qwen2} are an emerging paradigm for better intelligence in various speech and language tasks.
While most current research in Speech LLMs focuses on scaling up datasets and model parameters to enhance traditional tasks such as ASR, TTS, and emotion recognition, there has been limited exploration of their ability to handle broader conversational reasoning tasks that text-based LLMs excel at. Bridging this gap is critical for developing speech models that can engage in intelligent dialogue, without relying on intermediate text representations.

In this work, we propose a novel approach that leverages the ASR and TTS capabilities of Speech LLMs to enable natural speech conversations.
Our contributions can be summarized as follows:
\begin{itemize}
    \item We observed that direct finetuning of Speech LLMs on a medium scale of audio-only dataset yields incomprehensible speech conversation ability. Yet mixing the ASR transcripts as Chain of Thought (CoT) \citep{wei2022chain, zhang2023speechgpt} with input and output speech in the finetuning data yields better performance, at the cost of increased latency and data requirements. 
    \item To reduce the length of CoT tokens, we further propose to internalize ASR CoT tokens into a Speech LLM, retaining the speech conversation performance while reducing the latency by 14.5\%, and moving closer to a fully text-free end-to-end speech LLM.
    \item We constructed a large-scale synthetic conversational speech dataset with an emphasis on social common sense reasoning, containing $\sim$660k dialogue exchange pairs totaling $\sim$1000 hours of speech data. The dataset is publicly available on Huggingface Hub \footnote{\url{https://huggingface.co/datasets/robinysh/soda_tts}}. 
\end{itemize}

\vspace{-.5em}
\section{Background}
\vspace{-.5em}
\paragraph{Pretrained Speech and audio language models}
Since the success of scaling up LLMs, there are many attempts to scale up the speech models by training on discrete speech tokens \citep{lakhotia2021generative} or interleaving texts and speech inputs, such as SpeechT5 \citep{ao2022speecht5}, SpeechGPT series \citep{zhang2023speechgpt,zhang2024speechgpt,zhan2024anygpt} and Qwen-Audio series \citep{chu2023qwen,chu2024qwen2}.

SpeechGPT \citep{zhang2023speechgpt} and AnyGPT \citep{zhan2024anygpt} enable LLMs to perform speech understanding and generation through discrete speech representations.
Based on the pre-trained LLaMA2 7B model \citep{touvron2023llama}, they were further finetuned with large amounts of ASR and TTS datasets, and as a result, can perform well on those tasks.
However, attempts to perform direct speech-to-speech conversation with the model often result in incomprehensible audio.

\paragraph{Implicit Chain of Thought}
CoT prompting \citep{wei2022chain} is an effective method of improving LLMs' capability to perform complex reasoning tasks by detailing the intermediate steps. This technique has also been adapted to the multimodal domain, particularly in vision \citep{lu2022learn,zheng2023ddcot,mondal2024kam,zhangmultimodal}. Yet the application of CoT prompting in the audio and speech domain is still relatively rare \citep{li2024multimodal}.

Implicit Chain of Thought (ICoT), proposed by \cite{deng2024implicit}, has demonstrated LLMs can perform CoT behavior without explicit intermediate steps.
In \cite{deng2024explicitcotimplicitcot}, a curriculum learning approach was proposed to internalize the reasoning process by gradually removing the intermediate CoT tokens while retaining the reasoning performance.
The effectiveness of this approach was demonstrated in solving math problems.

\vspace{-.5em}
\section{Method}
\vspace{-.5em}
We incorporate ICoT \citep{deng2024implicit,deng2024explicitcotimplicitcot} to internalize ASR capability into a pre-trained Speech LLM, such that it performs speech-to-speech conversation without explicit ASR steps. 
To our knowledge, this is the first use of ICoT beyond math problems and into practical speech applications.
\paragraph{Base Model: AnyGPT}
We build on AnyGPT \citep{zhan2024anygpt},
an instruction-finetuned model capable of both speech audio comprehending and generation. AnyGPT is partly instruction-tuned on ASR and TTS tasks, resulting in strong capabilities in these areas. 
It is built upon the LLaMA 2 7B model \citep{touvron2023llama},
inheriting the strong zero-shot generalization capabilities that are characteristic of LLMs. 
This makes AnyGPT well-suited for our speech conversational tasks.

\begin{figure*}[]
\centering
\includegraphics[width=0.8\textwidth]{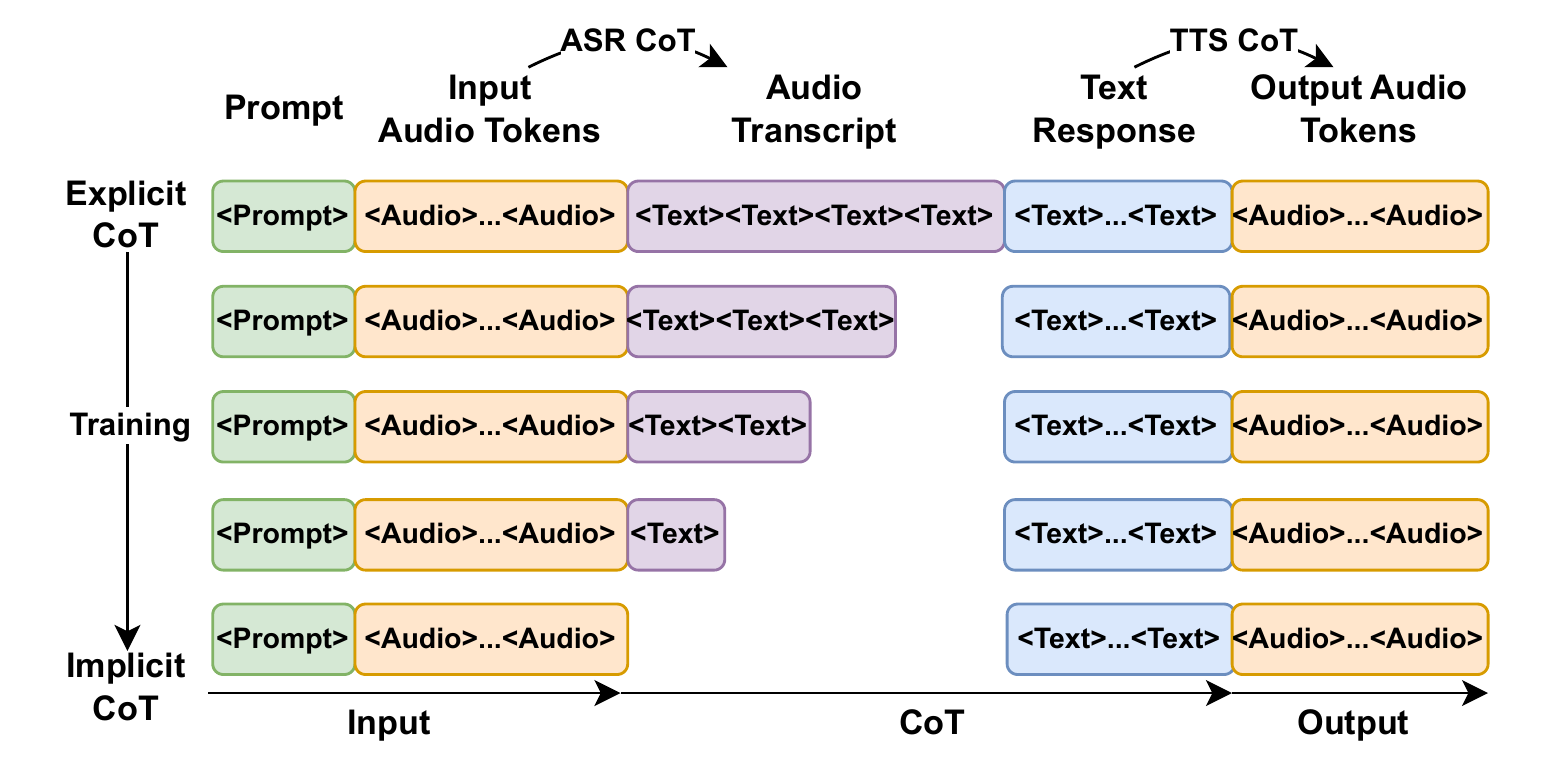}
\vspace{-1em}
\caption{Illustration of ICoT training and generation structure (from A-T-T-A to A-T-A ASR ICoT). Tokens of audio transcripts are removed linearly from the start during training, compressing the generation length for faster inference.}
\label{fig:icot}
\vspace{-1em}
\end{figure*}

\vspace{-.5em}
\paragraph{Speech-to-Speech Conversation via ASR and TTS CoT Prompting}
% Using AnyGPT’s ASR and TTS capabilities, we implement a CoT-based multi-step conversation framework similar to SpeechGPT \citep{zhang2023speechgpt}. Following \citet{xie2024miniomnilanguagemodelshear}, we use the "A-T-T-A" pipeline (Audio-Text-Text-Audio), where input audio is transcribed to text, a response is generated, and then converted to speech. The structured generation template, \textbf{"[CoT Prompt][Input Audio Tokens][Audio Transcript][Text Response][Output Audio Tokens]"}, is used to prompt each step. However, this intermediate text generation introduces latency and computation overhead.
Leveraging the ASR and TTS capability of AnyGPT,
speech-to-speech conversations can be implemented using a multi-step CoT strategy with intermediate text interfaces \citep{zhang2023speechgpt}. The conversational process is structured as follows: the model first transcribes the input audio via ASR, then generates a textual response, and finally converts the text into speech via TTS.
\citet{xie2024miniomnilanguagemodelshear} coined this pipeline as "\textbf{A-T-T-A}" (Audio-Text-Text-Audio), a terminlogy we adopt for clarity.
Additionally, we define the process of converting input audio tokens into audio transcript as \textbf{ASR CoT}, and the conversion of text response to output audio tokens as \textbf{TTS CoT}.

To induce zero-shot CoT behavior, we prepend a fixed CoT prompt at the start of the generation template. The full generation process can be roughly summarized with the template: \textbf{"[CoT Prompt][Input Audio Tokens][Audio Transcript][Text Response][Output Audio Tokens]"}. The exact template used can be found in Table-\ref{table:baseline_prompts}. However, this approach introduces two additional intermediate steps: the transcription of ASR output and the generation of a text-based response,
both introduce increased inference latency and computation overhead.

\paragraph{ASR Internalization via Implicit CoT (A-T-A ASR ICoT)}
To mitigate the latency caused by intermediate text generation, we adopt ICoT reasoning \cite{deng2024explicitcotimplicitcot}. This involves progressively internalizing the ASR reasoning steps during training, thereby eliminating the need for explicit transcription generation in the inference process, reducing the four-segment chain (\textbf{A-T-T-A}) to just three steps (\textbf{A-T-A}). Figure-\ref{fig:icot} illustrates this approach.
Training consists of two stages. \textbf{1)} Full-parameter fine-tuning of the model using standard CoT to ensure alignment with our prompts and dataset. \textbf{2)} ICoT training to internalize ASR CoT with LoRA \citep{hu2021loralowrankadaptationlarge}.
    
We define input audio tokens as $x$, intermediate audio transcript as $z^{ASR}=z^{ASR}_1,z^{ASR}_2,\cdots,z^{ASR}_m$, intermediate text response as $z^{TTS}=z^{TTS}_1,z^{TTS}_2,\cdots,z^{TTS}_n$, and final output audio tokens as $y$.
The language model with parameters $\theta$ is initially trained using a standard next-token prediction objective with the A-T-T-A format.
%The language model, with parameters $\theta$, is initially trained with the following objective:
%\begin{equation}
%    \min_\theta -\log P_\theta (y, z^{ASR}_{1:m},z^{TTS}_{1:n} \mid x) 
%\end{equation}
After this phase, we linearly reduce the number of audio transcript tokens to achieve ICoT. The number of tokens removed at each step $t$ is defined as:
\begin{equation}
   s(t) = \min\left( \left\lfloor \frac{t}{T} + o \right\rfloor, K_i \right)
\end{equation}
where $s(t)$ is the number of CoT tokens removed at step $t$,  $T$ is the number of steps per CoT token drop, $K_i$ is the amount of CoT tokens for data point $i$, and $o$ is a random variable sampled from an exponential distribution parameterized by $\lambda$. The new objective function becomes:
%where $s(t)$ is the number of CoT tokens removed at step $t$,  $T$ is the number of steps per CoT token drop, $K_i$ is the amount of CoT tokens for data point $i$, and $o$ is a random variable sampled from an exponential distribution parameterized by $\lambda$, i.e. $ P(o) \propto \exp(-\lambda o)$. The new objective function becomes:
\begin{equation}
    \min_\theta -\log P_\theta (y, z^{ASR}_{s(t):m},z^{TTS}_{1:n} \mid x) 
\end{equation}
\vspace{-1em}

Training continues until all audio transcript tokens are removed. To ensure stability during training, we also employed the optimizer reset strategy from \citet{deng2024explicitcotimplicitcot}.
%Every $T$ steps, the optimizer's state was reset, reinitializing both first and second-order gradients. This method prevents large estimates of second-order gradients resulting from a sudden removal of CoT tokens.  

\paragraph{Baselines}
To provide a meaningful comparison of our model's performance, we evaluate against the following baseline approaches. The exact prompt of the baselines can be found in Table-\ref{table:baseline_prompts}, and a summary of the differences between the baselines can be found in Table-\ref{table:baselines}.
\textbf{1) A-T-T-A Finetuned} To account for potential domain mismatches between the pre-training data and the target evaluation data, we fine-tune AnyGPT using our custom training prompts that include the ASR transcript.
%This model tests the effectiveness of domain-specific finetuning on model performance while keeping the ASR CoT mechanism intact.
\textbf{2) A-T-A no ASR ICoT} The model is fine-tuned directly on the conversational task without incorporating the ASR CoT mechanism. Only the text response is retained, and no ICoT training is applied. 
%This setup allows us to assess the impact of omitting the ASR ICoT training in favor of a more straightforward finetuning approach.
\textbf{3) A-A ICoT}
We attempted to internalize the TTS CoT step using ICoT.
%This explores the potential to extend the proposed method for training a fully text-less end-to-end speech LLM.
\textbf{4) A-T-T-A No Finetuning}
As a measure of the zero-shot capabilities of the base model, we evaluate AnyGPT's performance when prompted with an ASR CoT mechanism but without any task-specific finetuning.
%This model serves as a reference for the inherent capabilities of the pre-trained model in handling speech-to-speech tasks without additional domain adaptation.
\textbf{5) A-A No Finetuning} This assesses the capability of AnyGPT to understand and generate speech without CoT.
\textbf{6) Ground Truth Data} We utilize the ground truth test set from our TTS-generated SODA dataset.
%This oracle represents the expected upper bound for model performance.

%\begin{table}[ht!]
\begin{table}[]
    \parbox[t][][t]{.55\linewidth}{
    \setlength\tabcolsep{3pt}
    \resizebox{\linewidth}{!}{%
    \begin{tabular}{lrrc}
    \toprule
    \centering
    Model              & ASR prompt   & TTS prompt & Finetuned?\\
    \midrule
    A-T-T-A No Finetuning & Yes & Yes & \xmark \\ 
    A-A No Finetuning     & No & No & \xmark \\
    \midrule
    A-T-T-A Finetuned     & Yes & Yes & \cmark \\ 
    A-T-A No ASR ICoT     & No & Yes & \cmark \\ 
    \textbf{A-T-A ASR ICoT}  & Internalized & Yes & \cmark \\ 
    A-A ICoT & Internalized & Internalized & \cmark \\ 
    \bottomrule
    \end{tabular}
    }
    \caption{Summary of baseline models. "No" means the ASR/TTS prompt is removed even before training, whereas "internalized" means the prompt is being internalized via the ICoT process described in Fig.\ref{fig:icot}. Our proposed model is \textbf{bolded}.}
    \label{table:baselines}
    } 
    \hfill
    \parbox[t][][t]{.4\linewidth}{
    \centering
    \setlength\tabcolsep{3pt}
    \resizebox{\linewidth}{!}{%
    \begin{tabular}{lrr}
    \toprule
    \centering
                              & Train   & Test\\
    \midrule
    No. of dialogue pairs     & 663,103 & 6,540\\ 
    Total duration of audio   & 1098 hr & 10.7 hr\\ 
    Average duration of audio & 5.18 s  & 5.11 s\\ 
    WER                       & 9.00    & 9.30 \\ 
    No. of Unique Speakers    & 200000  & 2000 \\ 
    \bottomrule
    \end{tabular}
    }
    \caption{Statistics of synthetic SODA dataset}
    \label{table:dataset}
    }
\vspace{-1.5em}
\end{table}

\vspace{-.5em}
\section{Experiments}
\vspace{-.5em}
\paragraph{Dataset Construction}

Due to limited large-scale, publicly available speech-to-speech conversation datasets,
%Due to limited large-scale, open-source speech-to-speech conversational datasets, 
we synthesized TTS audio with SODA \citep{kim-etal-2023-soda},
a million-scale English dialogue dataset encompassing diverse social interactions.
Multi-turn dialogues were segmented into dialogue pairs and then converted into speech using \texttt{ChatTTS}\footnote{\url{https://github.com/2noise/ChatTTS}},
a TTS model for synthesizing high-quality conversational speech. 
To preserve speaker consistency,
unique speaker embeddings were sampled for each identity and maintained across dialogue pairs.

Following AnyGPT \citep{zhan2024anygpt},
we utilized \texttt{SpeechTokenizer}\footnote{\url{https://huggingface.co/fnlp/AnyGPT-speech-modules/tree/main/speechtokenizer}}\citep{zhang2023speechtokenizer} to tokenize each generated audio sample into discrete audio tokens.
This process resulted in 663,103 dialogue pairs, each represented as a tuple: \textbf{[}\textbf{Input Audio Tokens}, \textbf{Audio Transcript}, \textbf{Text Response}, \textbf{Output Audio Tokens}\textbf{]}.
%The structure represents each of the components in the \textbf{A-T-T-A} chain.
To verify audio quality, we employed \textit{Distill-Whisper-Large-V3}\footnote{\url{https://huggingface.co/distil-whisper/distil-large-v3}} \citep{gandhi2023distil} to calculate the Word Error Rate (WER). A summary of our dataset is presented in Table~\ref{table:dataset}.

\vspace{-.5em}
\paragraph{Training Setup}
All experiments used 4 Intel-Gaudi2 AI Accelerators and trained using AdamW \citep{loshchilov2018decoupled}.
Training was conducted in two stages.
The first stage involved standard CoT fine-tuning over 24,000 steps with a learning rate of 5e-6 and a batch size of 2 per device.
%In this stage, full fine-tuning on all parameters was adopted to help the model learn from conversational datasets.  
For the second stage, we used LoRA \citep{hu2021loralowrankadaptationlarge} for another 24,000 steps, increasing the batch size to 4 and the learning rate to 5e-5.
The LoRA was integrated into the attention mechanisms, using a rank of 32 and an alpha value of 32.
To implement ICoT reasoning, we progressively removed one audio transcription token every $T=500$ steps, with removal smoothing parameter $\lambda=4$.
Additionally, a third stage was attempted for TTS ICoT, in which we removed one text response token every $T=2000$ steps, and applied a smaller learning rate of 2e-6 to ensure stability.

%\section{Evaluation}
%\paragraph{LLM Evaluation}
\vspace{-.5em}
\paragraph{Evaluation}
%Given the high cost of human evaluation, recent research has adopted a new paradigm that uses strong language models like GPT-4 and ChatGPT for automatic scoring or comparison with human responses \citep{zheng2023judging,rafailov2023direct,dubois2024lengthcontrolled}.
We evaluate the model's ability to accurately understand input audio and generate coherent,
contextually appropriate responses using \textbf{Prometheus-Eval 2.7B} \citep{kim2024prometheus}, an LLM explicitly fine-tuned for evaluating text attributes with customizable metrics, alongside with \textbf{GPT-4o}. We employed two evaluation models to minimize bias from any single model.
Responses were scored on two metrics: \textbf{Naturalness} (fluency and human-likeness), and \textbf{Specificity} (relevance and contextual alignment).
Evaluation prompts were manually crafted, and the evaluator compared outputs to determine win rates. The prompts can be found in Table-\ref{table:eval_prompt} and Table-\ref{table:gpt4o_eval_prompt}.
%Since Prometheus-Eval and GPT-4o are text-based, they cannot directly evaluate speech outputs.
Since Prometheus-Eval and GPT-4o are text-based, 
we transcribed all generated audio via \textbf{Distill-Whisper-Large-V3}, before feeding into the evaluators.
Win rates against our model are shown in Fig \ref{fig:winrate_vs_model}, while comparisons against the ground truth provided in the appendix \ref{appendix:eval_groundtruth}.

To validate the use of LLM evaluation as a proxy for human judgment, two authors blind-tested baseline models against A-T-A (ASR ICoT) samples.
The Cohen’s Kappa between human evaluators and GPT-4o's evaluations was \textbf{0.586}, demonstrating reasonable consistency between LLM evaluations and human evaluations.

\begin{figure}[]
\centering
\includegraphics[width=1.1\textwidth, center]{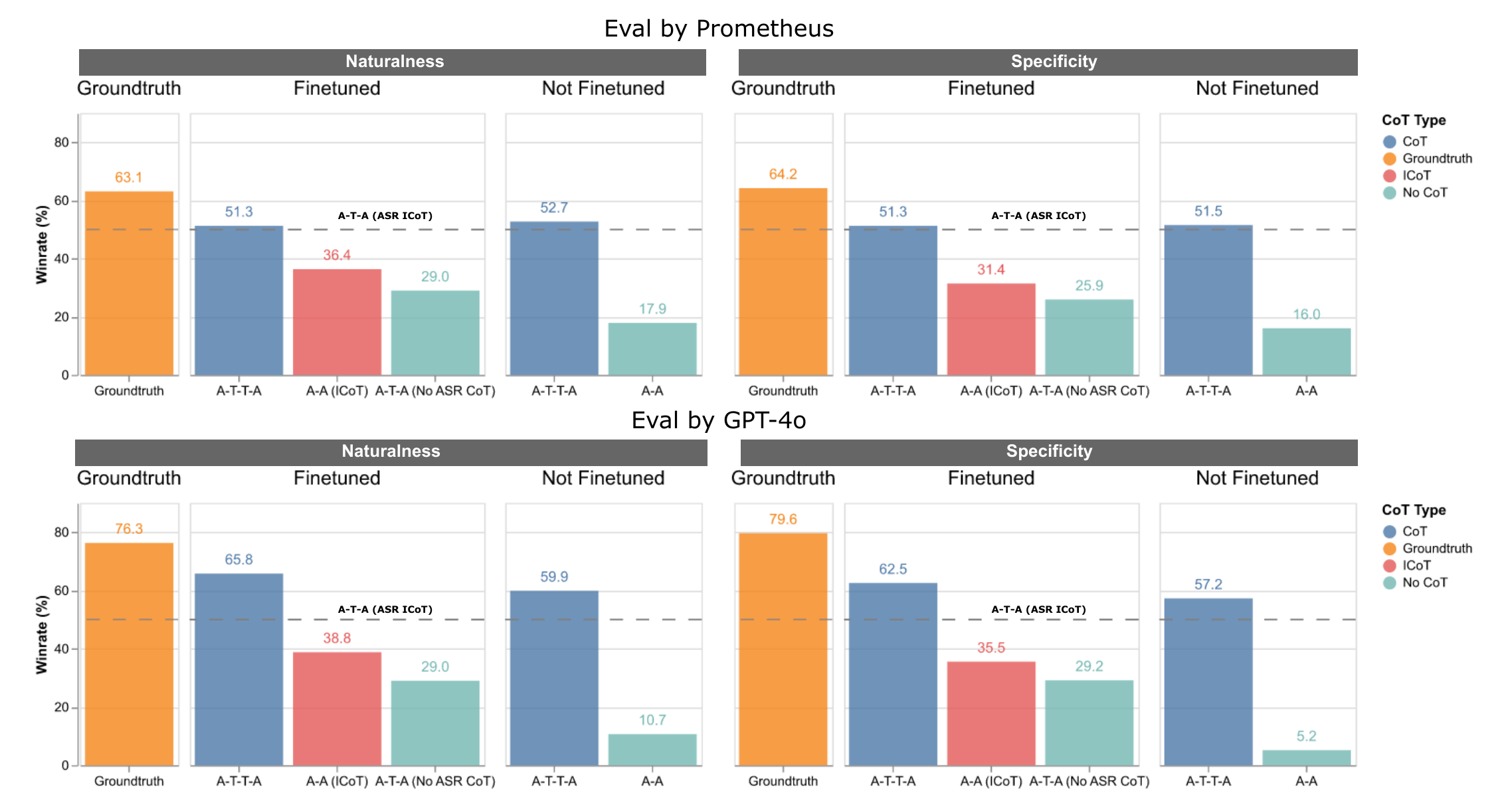}
\vspace{-2em}
\caption{Winrate (percentage) of different models’ generated responses compared to the proposed \textbf{A-T-A (ASR ICoT)} model, as evaluated by Prometheus and GPT-4o.
A higher percentage indicates that the model outperformed the proposed method more often.
The dotted line marks a 50\% winrate (draw).
While our model did not surpass the slower A-T-T-A chain-of-thought method, it outperformed most baseline models significantly, particularly those with similar or lower latency.}
\label{fig:winrate_vs_model}
\vspace{-1em}
\end{figure}

\vspace{-.5em}
\paragraph{Latency} We measured the inference latency on an Nvidia 3090 GPU using Huggingface Transformers with KV-cache and FlashAttention 2 \cite{dao2023flashattention2}.
To simulate a streaming setting, latency was measured from when the model received the last audio input token to the generation of the first output audio token. The results are presented in Table-\ref{table:latency}.

\begin{table}[]
    \centering
    \small
    \begin{tabular}{@{}lrr@{}}
    \toprule
    \centering
    \textbf{Model} & \textbf{A-T-A (ASR ICoT)} & \textbf{A-T-T-A (Fineutued)}\\
    \midrule
    \textbf{Latency (s)} ($\downarrow$)     & \textbf{0.87} & 1.09  \\ 
    % \quad $-$  $\mathcal{L}_{\text{distill}}$ & TBA &  TBA  \\ 
    \textbf{Mean Generated Audio Transcript Count}  & 0.0 & 21.3\\ 
    \textbf{Mean Generated Text Response Count}  & 19.6 & 22.1 \\ 
    \bottomrule
    \end{tabular}
    \caption{Inference Statistics. The best latency is bolded.}
    \label{table:latency}
\vspace{-2em}
\end{table}

\vspace{-.5em}
\section{Results} 
\vspace{-.5em}
%In this section, we summarize the main findings of our experiments.
%A selection of transcribed generated samples is provided in Table-\ref{table:dialogue_samples}.

\paragraph{ICoT Effectively Internalizes ASR Capabilities}
As shown in Figure-\ref{fig:winrate_vs_model}, our ICoT-trained model internalizes ASR effectively.
When compared to the \textbf{A-T-T-A} finetuned model with explicit ASR CoT, our model achieves a competitive win rate of \textbf{42.3\%}, averaged over each LLM evaluator and metrics.
This suggests that internalizing ASR CoT introduces only minor quality degradation.

In contrast, the \textbf{A-T-A (No ASR CoT)} model trained with direct finetuning without any ASR CoT shows a significant drop in performance.
Our model outperforms significantly with a win rate of \textbf{71.7\%}, averaged over each LLM evaluator and metrics.
As shown in Table-\ref{table:dialogue_samples}, the samples generated by \textbf{A-T-A (No ASR CoT)} are grammatically incoherent.
%Directly bypassing the ASR transcripts in finetuning does not work effectively, as the model still relies heavily on transcripts in reasoning the contents of its response.
These results underscore the necessity of ICoT in preserving the quality of speech-to-speech interactions while eliminating the need for explicit ASR steps.

\vspace{-.5em}
\paragraph{Internalizing ASR Reduces Latency}
In addition to maintaining competitive conversational quality, internalizing the ASR process leads to notable efficiency gains, as fewer tokens are needed in inference.
By removing the ASR CoT, the latency for generating the first output audio token with KV cache decreased from 1.09 seconds to 0.87 seconds, a relative reduction of 20.2\%.

\vspace{-.5em}
\paragraph{ICoT Does Not Fully Generalize to TTS Internalization}
As shown in Figure \ref{fig:winrate_vs_model}, applying ICoT to internalize the TTS process results in a notable decline in both \textbf{Naturalness} and \textbf{Specificity}. Unlike \textbf{A-T-A (ASR ICoT)}, the win rate for \textbf{A-A (ICoT)} drops significantly, averaging only \textbf{35.5\%}. The model struggles to generate contextually relevant speech, as illustrated by the example in Table \ref{table:dialogue_samples}.
These results suggest that while ICoT is effective for ASR, further research is needed to refine its application for TTS, where explicit textual processing appears crucial for maintaining high-quality audio responses \cite{xie2024miniomnilanguagemodelshear, fang2024llamaomniseamlessspeechinteraction,defossez2024moshi}.

\vspace{-.5em}
\section{Conclusion}
\vspace{-.5em}
We presented a method for internalizing the ASR CoT in large Speech LLMs, enabling speech-to-speech conversations without explicit ASR steps. Our approach reduces the inference latency of Speech LLMs while maintaining high-quality conversational performance through ICoT reasoning.
Additionally, we contributed a large-scale synthetic conversational speech dataset and introduced an evaluation pipeline using LLMs for scoring naturalness and specificity.
%Future research could focus on refining these techniques and improving evaluation methods for speech-to-speech models.

\begin{ack}
This research was enabled in part by support provided by Advanced Research Computing at the University of British Columbia, the Digital Research Alliance of Canada, LAION and Intel through LAION's BUD-E project.
We acknowledge the support of the Discovery Program from the Natural Sciences and Engineering Research Council of Canada (NSERC).
\end{ack}

\bibliography{refs}

\newpage
\appendix

\section{Appendix / supplemental material}

\subsection{Evaluation results against groundtruth}
\label{appendix:eval_groundtruth}
We report the winrate of our proposed model and baselines against the groundtruth evaluated by Prometheus and GPT-4o. The results can be found in Figure-\ref{appendix:eval_groundtruth}. They show a similar trend to Figure-\ref{fig:winrate_vs_model}, indicating that both of our evaluation models are consistent with their scoring. 

Additionally, the authors conducted a blind trial comparing ground truth samples with outputs from the proposed models and baselines. The trial yielded a Cohen's Kappa score of \textbf{0.389} when compared with GPT-4o's evaluations. The relatively lower agreement score is attributed to the ambiguity in cases where samples were of comparable quality to the ground truth, making the choice less definitive.

\begin{figure*}[ht!]
\centering
\includegraphics[width=1.1\textwidth, center]{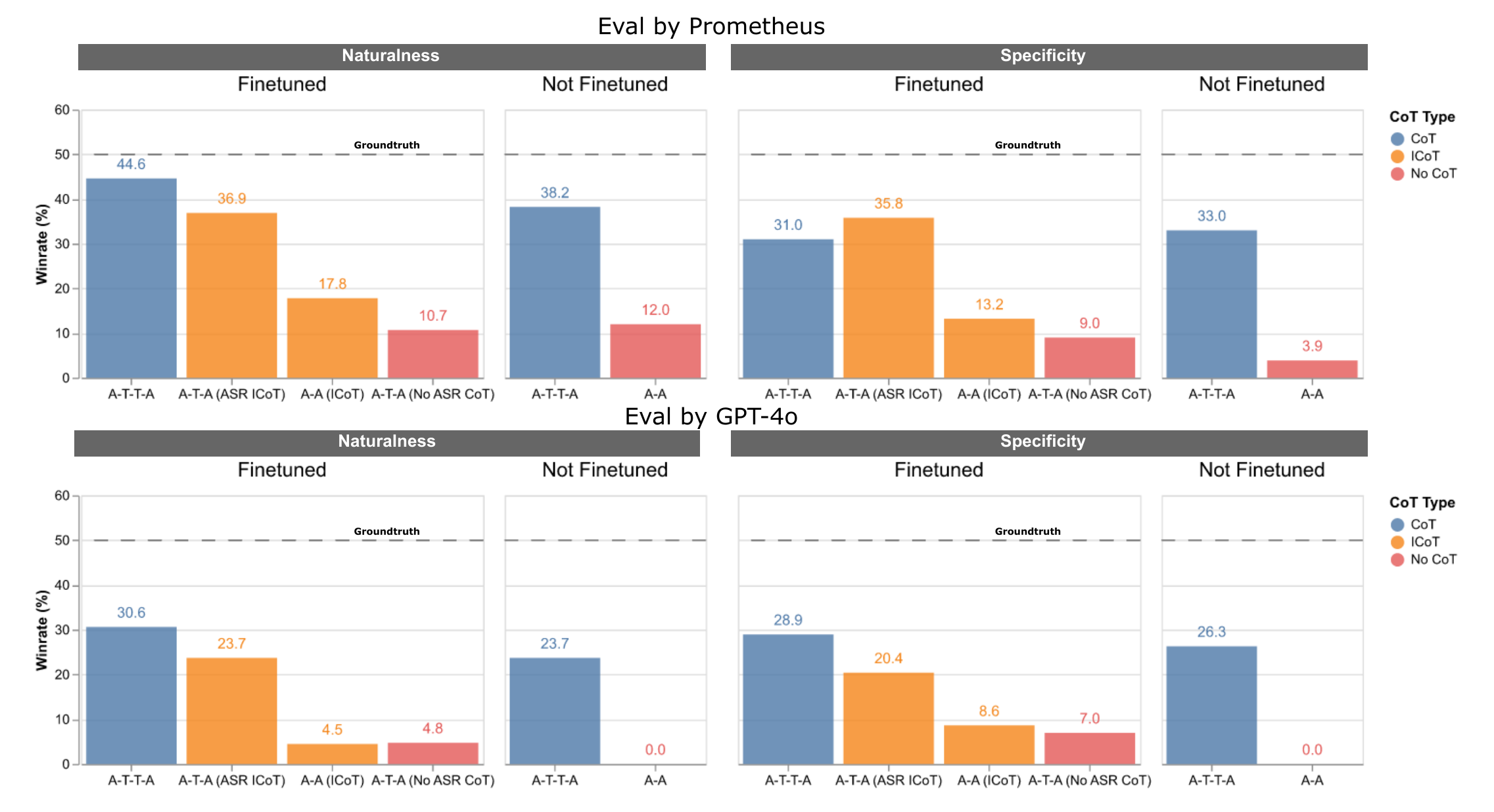}
\vspace{-2em}
\caption{Winrate (percentage) of different models’ generated responses compared to the groundtruth, as evaluated by Prometheus and GPT-4o.
A higher percentage indicates that the model outperformed the groundtruth more often.
The dotted line marks a 50\% winrate (draw).
The results show consistent trends compared with Figure-\ref{fig:winrate_vs_model}.
}
\label{fig:winrate_vs_gt}
\end{figure*}

\newpage
\subsection{Prompts used in evaluation}
Table~\ref{table:eval_prompt} contains the evaluation criteria that we used to prompt Prometheus \citep{kim2024prometheus}. Table~\ref{table:gpt4o_eval_prompt} shows the prompts for the GPT-4o evaluation.  
\begin{table}[h]%[H]
    \centering
    %\small
    %\vspace{1ex}
    \begin{tabular}{p{.1\textwidth}p{.4\textwidth}p{.4\textwidth}}
    \hline
    %\centering
              & \textbf{Evaluation of Naturalness} & \textbf{Evaluation of Specificity} \\ 
    \hline
    \textbf{Criteria} & How smooth, fluid, and human-like the response sounds, without awkward phrasing or robotic tone. & How closely the response is tailored to the preceding message, ensuring it directly addresses the context and intent with relevant details. \\ 
    \hline
    \multicolumn{3}{c}{Scoring Rubrics}\\
    \hline
    \textbf{score=1}                &  The response is highly robotic, awkward, or stilted. It feels forced and does not resemble natural human speech. The grammar and phrasing may be incorrect, and the response does not flow smoothly.  & The response is highly generic and does not address the context or intent of the conversation. It feels like a random, unrelated statement that does not meaningfully connect to the previous message. \\
    \hline
    \textbf{score=2}                            &  The response is somewhat awkward or lacks fluidity. While the sentence structure may be understandable, the conversation feels rigid, with obvious flaws in phrasing and tone. It doesn't sound like how a human would naturally speak.  & The response is somewhat related to the context but remains too vague or generic. While it acknowledges the previous message, it lacks detail and does not directly engage with the specific content or intent of the dialogue. \\
    \hline
    \textbf{score=3}                            &  The response is generally understandable and flows reasonably well. However, there are still noticeable unnatural patterns or awkward phrasing that make it feel somewhat artificial. It might pass as natural in some instances, but not consistently.  & The response addresses the context in a general way. While it does connect to the previous message, it still lacks deeper engagement or precision. It answers at a surface level without delving into specific details. \\[10pt]
    \hline
    \textbf{score=4}                            &  The response flows well and closely resembles natural human conversation. While there may be minor imperfections or slightly formal language, it feels smooth and engaging, with little to no awkwardness.  & The response is tailored to the context and shows a clear understanding of the previous message. It includes relevant details and addresses the main points of the conversation, although there might be minor areas where it could be more specific. \\
    \hline
    \textbf{score=5}                            &  The response feels entirely fluid and natural, as if it were generated by a human speaker. The tone, phrasing, and sentence structure are perfect, with no signs of robotic or awkward language.  & The response is fully tailored to the context, addressing the previous message in a highly relevant and detailed manner. It demonstrates a clear and precise understanding of the conversation, engaging deeply with all the important elements. \\
    \hline
    \end{tabular}
    %\vspace{-0.3cm}
    \caption{Prompts used for Prometheus evaluation}
    \label{table:eval_prompt}
\end{table}

%\subsection{Table of prompts used in evaluation by GPT-4o}
\begin{table}[h]
    \centering
    \small
    \resizebox{\linewidth}{!}{%
    \begin{tabular}{@{}lp{.9\linewidth}}
    \toprule
    \textbf{Rubric} & \textbf{Prompt}\\ 
    \midrule
    Naturalness&                     You are a fair judge assistant assigned to deliver insightful feedback that compares individual performances, highlighting how each stands relative to others within the same cohort.
    
    \#\#\# Task Description:
    You will evaluate the quality of two responses to a dialogue input snippet from a larger dialogue. Responses to evaluate, and a score rubric representing an evaluation criteria are given.
    
    1. Write a detailed feedback that assesses the quality of two responses strictly based on the given score rubric, not evaluating in general.
    
    2. After writing feedback, choose a better response between Response A and Response B. You should refer to the score rubric.
    
    3. The output JSON format should look as follows:
    \{
        'explanation': 'Write a feedback for each response and give your explanation for the choice',
        'winner': 'A' or 'B'
    \}

    \#\#\# Score Rubric:
    
    [Naturalness: How smooth, fluid, and human-like the response sounds, without awkward phrasing or robotic tone.]
    
    - Score 1: The response is highly robotic, awkward, or stilted. It feels forced and does not resemble natural human speech. The grammar and phrasing may be incorrect, and the response does not flow smoothly.
      
    - Score 2: The response is somewhat awkward or lacks fluidity. While the sentence structure may be understandable, the conversation feels rigid, with obvious flaws in phrasing and tone. It doesn't sound like how a human would naturally speak.
      
    - Score 3: The response is generally understandable and flows reasonably well. However, there are still noticeable unnatural patterns or awkward phrasing that make it feel somewhat artificial. It might pass as natural in some instances, but not consistently.
      
    - Score 4: The response flows well and closely resembles natural human conversation. While there may be minor imperfections or slightly formal language, it feels smooth and engaging, with little to no awkwardness.
      
    - Score 5: The response feels entirely fluid and natural, as if it were generated by a human speaker. The tone, phrasing, and sentence structure are perfect, with no signs of robotic or awkward language.
    
    \#\#\# Dialogue Input:
    \textbf{<Text transcript of input audio tokens from the dataset>}
    
    \#\#\# Response A:
    \textbf{<Whisper ASR transcript of audio response A>}
    
    \#\#\# Response B:
    \textbf{<Whisper ASR transcript of audio response B>}
    
    \#\#\# Feedback:
    \\
    \midrule
     Specificity &                     You are a fair judge assistant assigned to deliver insightful feedback that compares individual performances, highlighting how each stands relative to others within the same cohort.

\#\#\# Task Description:
You will evaluate the quality of two responses to a dialogue input snippet from a larger dialogue. Responses to evaluate, and a score rubric representing an evaluation criteria are given.

1. Write a detailed feedback that assesses the quality of two responses strictly based on the given score rubric, not evaluating in general.

2. After writing feedback, choose a better response between Response A and Response B. You should refer to the score rubric.

3. The output JSON format should look as follows:
\{
    'explanation': 'Write a feedback for each response and give your explanation for the choice',
    'winner': 'A' or 'B'
\}

\#\#\# Score Rubric:

[Specificity: How closely the response is tailored to the preceding message, ensuring it directly addresses the context and intent with relevant details.] 

- Score 1: The response is highly generic and does not address the context or intent of the conversation. It feels like a random, unrelated statement that does not meaningfully connect to the previous message.
  
- Score 2: The response is somewhat related to the context but remains too vague or generic. While it acknowledges the previous message, it lacks detail and does not directly engage with the specific content or intent of the dialogue.
  
- Score 3: The response feels entirely fluid and natural, as if it were generated by a human speaker. The tone, phrasing, and sentence structure are perfect, with no signs of robotic or awkward language.
  
- Score 4: The response is tailored to the context and shows a clear understanding of the previous message. It includes relevant details and addresses the main points of the conversation, although there might be minor areas where it could be more specific. 
  
- Score 5: The response is fully tailored to the context, addressing the previous message in a highly relevant and detailed manner. It demonstrates a clear and precise understanding of the conversation, engaging deeply with all the important elements.

\#\#\# Dialogue Input:
\textbf{<Text transcript of input audio tokens from the dataset>}

\#\#\# Response A:
\textbf{<Whisper ASR transcript of audio response A>}

\#\#\# Response B:
\textbf{<Whisper ASR transcript of audio response B>}

\#\#\# Feedback: \\
    \bottomrule
    \end{tabular}
    }
    %\vspace{-0.3cm}
    %\caption{List of prompts for different models.}
    \caption{Prompts used for GPT-4o Evaluation}
    \label{table:gpt4o_eval_prompt}
\end{table}

\clearpage
\subsection{Prompts for training and inference}
Table \ref{table:baseline_prompts} contains the prompt used for training and inferencing each of the baselines and proposed model.
\begin{table}[h]
    \centering
    \small
    \vspace{1ex}
    \begin{tabular}{@{}lp{0.65\linewidth}}
    \toprule
    \centering
    \textbf{Model} & \textbf{Prompt}\\ 
    \midrule
    A-T-T-A (Finetuned/Not Finetuned)& \texttt{You are [AnyGPT]. You are chatting with [Human]. Step by step, give me the transcript of the provided audio, a chat response to the transcript, and read the response. <-Ins-> [Human]: \textbf{<Input Audio Tokens>}<eoh> [AnyGPT]: <-Res-> \textbf{<Audio Transcript>}\textbackslash n[AnyGPT]: \textbf{<Text Response>} \textbf{<Output Audio Tokens>}<eos>} \\ 
    \midrule
    A-T-A (ASR ICoT/No ASR CoT) & \texttt{You are [AnyGPT]. You are chatting with [Human]. Step by step, give me the transcript of the provided audio, a chat response to the transcript, and read the response. <-Ins-> [Human]: \textbf{<Input Audio Tokens>}<eoh> [AnyGPT]: <-Res-> [AnyGPT]: \textbf{<Text Response>} \textbf{<Output Audio Tokens>}<eos>} \\ 
    \midrule
    A-A (ICoT) & \texttt{You are [AnyGPT]. You are chatting with [Human]. Step by step, give me the transcript of the provided audio, a chat response to the transcript, and read the response. <-Ins-> [Human]: \textbf{<Input Audio Tokens>}<eoh> [AnyGPT]: <-Res-> [AnyGPT]: \textbf{<Output Audio Tokens>}<eos>} \\ 
    \midrule
    A-A (Not Finetuned) & \texttt{You are [AnyGPT]. You are chatting with [Human]. Give me a speech response to [Human]. <-Ins-> [Human]: \textbf{<Input Audio Tokens>}<eoh> [AnyGPT]: <-Res->\textbf{<Output Audio Tokens>}<eos>} \\ 
    \bottomrule
    \end{tabular}
    %\vspace{-0.3cm}
    %\caption{List of prompts for different models.}
    \caption{List of prompts for different models. \textbf{<Input Audio Tokens>} represents the input audio tokens generated by SpeechTokenizer, \textbf{<Audio Transcript>} denotes the audio transcript of the input audio, \textbf{<Text Response>} denotes the output text response, and \textbf{<Output Audio Tokens>} denotes the output Audio Tokens.}
    \label{table:baseline_prompts}
\end{table}

\clearpage
\subsection{ASR Transcribed model output samples}
Several samples generated by the baselines and our proposed model were transcribed by \textbf{Distill-Whisper-Large-V3}\citep{gandhi2023distil} and presented in Table \ref{table:dialogue_samples}.
\begin{table}[h]
    \centering
    \small
    \begin{tabular}{@{}lp{0.65\linewidth}}
    \toprule
    \multicolumn{2}{c}{Case 1}\\
    \midrule
    \textbf{Dialogue Input}& I love the snow! It's so beautiful and peaceful.\\
    \midrule
    \textbf{Ground-truth response}& It is pretty, but I don't like how cold it is.\\
    \midrule
    \textbf{A-T-A* (ASR ICoT, Finetuned)}& It is indeed. It has a calming effect. \\
    \textbf{A-A (ICoT, Finetuned)}& That's such a great plan. I don't know what I want.\\
    \midrule
    \textbf{A-T-A (No ASR CoT, Finetuned)}& So what do you have any different standards i think it's probably safe on on difficult intentity in explaining any speakin\\
    \textbf{A-T-T-A (Finetuned)}& I can imagine it's a really good book. \\
    \midrule
    \textbf{A-A (Not Finetuned)}&   savant seventh and peace for self and peace for self and peace for self our posidancy having dover rossett in seven and seven and seven and seven (...repeating)\\
    \textbf{A-T-T-A (Not Finetuned)}& I can understand why you like it. It's a very calming song.\\
    \midrule
    \multicolumn{2}{c}{Case 2}\\
    \midrule
    \textbf{Dialogue Input}& So what have you been up to lately?\\
    \midrule
    \textbf{Ground-truth response}& Not much, just hanging out with family and friends mostly. You?\\
    \midrule
    \textbf{A-T-A* (ASR ICoT, Finetuned)}& I've been up to absolutely nothing today. How about you? \\
    \textbf{A-A (ICoT, Finetuned)}& I'm coming. Thanks for doing some work with.\\
    \midrule
    \textbf{A-T-A (No ASR CoT, Finetuned)}& No, no, no, I don't know happy for my family and friends done well.\\
    \textbf{A-T-T-A (Finetuned)}&  just hanging out with friends going to school and work. You know, the usual.\\
    \midrule
    \textbf{A-A (Not Finetuned)}& see whide's a baby times of cordrate you're gonna fissionate in the second cullinose deservin ah a one o times a bring one wonderful twey j\\
    \textbf{A-T-T-A (Not Finetuned)}& I have been up to my daily routine.\\
    \midrule
    \end{tabular}
    \caption{Whisper ASR-transcribed inputs and responses from different models compared with the ground-truth response. The model with \textbf{*} is our proposed method. The results from \textbf{A-T-A (No ASR CoT, Finetuned)} and \textbf{A-A (Not Finetuned)} reveal that the absence of ASR CoT or ICoT leads to grammatically incoherent speech. In contrast, \textbf{A-A (ICoT, Finetuned)} produces grammatically coherent but contextually irrelevant speech.}
    \label{table:dialogue_samples}
\end{table}

\end{document}